\documentclass[letterpaper, 10 pt, journal, twoside]{ieeetran} 
\IEEEoverridecommandlockouts                             

\usepackage{graphics}

\usepackage{amsmath}
\usepackage{amssymb}  
\usepackage{siunitx}
\usepackage{mathtools}
\usepackage{booktabs}
\usepackage[ruled]{algorithm2e}
\usepackage{subcaption}
\usepackage[bottom=57pt,top=54pt, left=54pt, right=54pt]{geometry}   %57-43

\usepackage{adjustbox}
\usepackage{pbox}
\usepackage{dblfloatfix}
\usepackage{graphicx}
\usepackage[dvipsnames]{xcolor}
\usepackage{hyperref}
\usepackage{caption}
\usepackage{subcaption}
\usepackage[nospace]{cite}

\usepackage{array,multirow,graphicx}
\usepackage{makecell}
\usepackage{pifont}
\usepackage{xparse}
\usepackage{enumitem}
\usepackage{float}
\floatstyle{plaintop}
\restylefloat{table}
\usepackage{arydshln}
\usepackage[normalem]{ulem}
\usepackage{pifont}

\NewDocumentCommand{\todo}{o m}{\textcolor{red}{\textbf{TODO\IfNoValueTF{#1}{}{(#1)}:} #2}}
\NewDocumentCommand{\note}{o m}{\textcolor{orange}{\textbf{NOTE\IfNoValueTF{#1}{}{(#1)}:} #2}}
\newlist{todolist}{itemize}{2} \setlist[todolist]{label=$\square$}

\newcommand{\etal}{\textit{et~al.~}}
\newcommand{\rev}[1]{#1}  %\textcolor{blue}{#1}
\newcommand{\quali}[2]{\textbf{\textcolor[HTML]{#1}{#2}}}

\begin{document}

\begin{minipage}{0.8\textwidth} \copyright 2023 IEEE. Personal use of this material is permitted. Permission from IEEE must be obtained for all other uses, in any current or future media, including reprinting/republishing this material for advertising or promotional purposes, creating new collective works, for resale or redistribution to servers or lists, or reuse of any copyrighted component of this work in other works.\\ Please cite this paper as:\\
\begin{verbatim} 
@article{
  Schmid23Dynablox, 
  author={Schmid, Lukas and Andersson, Olov and Sulser, Aurelio and
          Pfreundschuh, Patrick and Siegwart, Roland},
  journal={IEEE Robotics and Automation Letters}, 
  title={Dynablox: Real-Time Detection of Diverse Dynamic Objects in
         Complex Environments}, 
  year={2023},
  volume={8},
  number={10},
  pages={6259-6266},
  doi={10.1109/LRA.2023.3305239}}
} \end{verbatim} \end{minipage}
\newpage

\title{Dynablox: Real-time Detection of Diverse Dynamic Objects in Complex Environments}

\markboth{IEEE Robotics and Automation Letters. Preprint Version. Accepted August, 2023.}
{Schmid \MakeLowercase{\textit{et al.}}: Dynablox: Real-time Detection of Diverse Dynamic Objects in Complex Environments}  

\author{Lukas Schmid$^{1,2\ast}$, Olov Andersson$^{1\ast}$, Aurelio Sulser$^1$, Patrick Pfreundschuh$^1$, and Roland Siegwart$^1$% <-this % stops a space

\thanks{Manuscript received: April, 20, 2023; Revised July, 11, 2023; Accepted August, 4, 2023.}%Use only for final RAL version
\thanks{This paper was recommended for publication by Editor Editor J. Civera upon evaluation of the Associate Editor and Reviewers' comments.}%
\thanks{$^1$Autonomous Systems Lab, ETH Z\"urich, Switzerland. \newline {\tt\footnotesize \{nandersson, asulser, patripfr, rsiegwart\}@ethz.ch}}%
\thanks{$^2$Massachusetts Institute of Technology, MA, USA. {\tt\footnotesize lschmid@mit.edu}}%
\thanks{$^\ast$Indicates equal contribution.}%
\thanks{This work has received funding from the Microsoft Swiss Joint Research Center, the Swiss National Science Foundation (SNSF), a Wallenberg Foundation and WASP Postdoctoral Scholarship, and the Swiss National Science Foundation's NCCR DFab P3.}%
\thanks{Digital Object Identifier (DOI): see top of this page.}%
}%

\maketitle

%%%%%%%%%%%%%%%%%%%%%%%%%%%%%%%%%%%%%%%%%%%%%%%%%%%%%%%%%%%%%%%%%%%%%%%%%%%%%%%%
\begin{abstract}
Real-time detection of moving objects is an essential capability for robots acting autonomously in dynamic environments.
We thus propose \emph{Dynablox}, a novel online mapping-based approach for robust moving object detection in complex \rev{unstructured} environments.
The central idea of our approach is to incrementally estimate high confidence free-space areas by modeling and accounting for sensing, state estimation, and mapping limitations during online robot operation.
The spatio-temporally conservative free space estimate enables robust detection of moving objects without making any assumptions on the appearance of objects or environments.
This allows deployment in complex scenes such as multi-storied buildings or staircases, and for diverse moving objects such as people carrying various items, doors swinging or even balls rolling around. 
We thoroughly evaluate our approach on real-world data sets, achieving 86\% IoU at 17 FPS in typical robotic settings.
The method outperforms a recent appearance-based classifier and approaches the performance of offline methods. 
We demonstrate its generality on a novel data set with rare moving objects in complex environments. 
We make our efficient implementation and the novel data set available as open-source.
\end{abstract}

\vspace{-10pt}

% ===============================================================================================
\vspace{-12pt}

\section{Introduction}
\label{sec:introduction}
\begin{figure}[t]
    \centering
    \includegraphics[width=\linewidth]{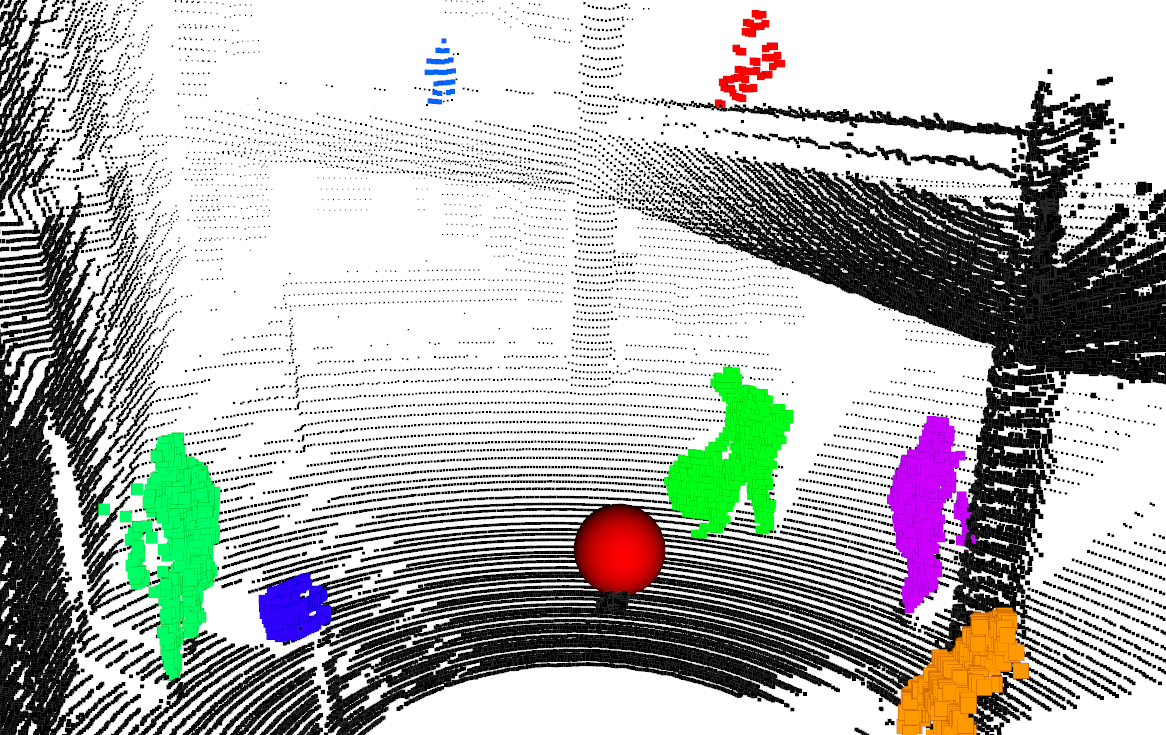}
    \caption{Dynablox accurately detects diverse moving objects such as people and the rolling (blue) and carried (green) balls in a complex two-story environment during online operation (sensor shown red).}
    \label{fig:teaser}
    \vspace{-20pt}
\end{figure}

\IEEEPARstart{A}{s} the capabilities of mobile robots are advancing beyond controlled environments to diverse tasks such as inspection, security, or logistics in complex dynamic environments, the need to reliably detect a variety of moving objects such as people, vehicles, and other robots is of utmost importance for safe robot operation. 
This is a challenging problem for several reasons.
In many cases, autonomous robots have to act in new or rapidly changing environments. 
They thus must detect moving objects online and on-board instead of relying on an existing map. 
In addition, real-world workplaces and public spaces come with multiple challenges in that they do not exhibit strong structure such as flat streets or marked pedestrian crossings. 
Instead there is large variety in environment structure, such as tunnels in mines to shopping malls with open floor plans in multiple levels, stairs, or ramps, as well as in the type of moving objects encountered, such as children playing ball, pets, or people carrying objects.
However, most work on online moving object detection make heavy use of appearance-based cues, either engineered for a particular type of environment, or trained offline using labeled examples. 
This makes them vulnerable to failing on uncommon objects or non-planar environments such as in Fig.~\ref{fig:teaser}. 

To enable robust moving object detection over heterogeneous unstructured environments and objects, we here instead propose an online mapping-based approach that is agnostic to appearance by extracting motion cues from state-of-the-art solutions for volumetric mapping.
Our approach is based on the insight that to robustly detect moving objects using a map, we need a robust notion of free space. 
However, this is particularly challenging when the map is constructed online, as errors arise not only from sensor noise, but also state drift, map uncertainty at the boundaries of unexplored space, as well as sparse returns at range. 
We detail in turn how to overcome these issues to estimate the set of voxels that are free with high confidence.
These are then used to recover all moving points, and update the map accordingly.
Our method thus does not require advance knowledge of the type of environment or moving object that the robot will encounter. 
Second, it adds little computational overhead over a conventional mapping stack required for autonomous navigation in such environments, allowing it to run online and in real-time on-board a laptop-grade CPU.  

While the method is general, in this work we focus on robots equipped with a LiDAR sensor. 
LiDAR has emerged as a reliable choice for mapping complex unknown environments in challenging conditions, with teams in the recent DARPA Subterannean Challenge relying heavily on it \cite{tranzatto2022team,ramezani2022wildcat,palieri2020locus,zhao2021super}. 
As there is still a lack of open-source solutions for 3D moving object detection in unstructured environments, we make our implementation available as a layer running on top of the popular mapping framework Voxblox \cite{oleynikova2017voxblox}, which was also used by the DARPA SubT winning team\cite{tranzatto2022team}.

We make the following contributions: 
\begin{itemize}
    \item We propose a methodology for incremental high-confidence free space estimation from point cloud data, modeling sensor noise, measurement sparsity, dynamic environments, and state estimation drift.
    \item We present \emph{Dynablox}, our system leveraging high-confidence free space for robust real-time detection of diverse dynamic objects in complex \rev{unstructured} environments.
    \item We thoroughly evaluate the presented approach, achieving 86\% IoU at 17 FPS on a mobile CPU, and demonstrate its robustness to radically different object types, diverse environments, and drifting state estimates. We make our efficient implementation and the newly recorded data available as open-source\footnote{Released at \url{https://github.com/ethz-asl/dynablox}.}.
     
\end{itemize}
% The remainder of the paper is structured as follows. We describe related work in section~\ref{sec:rel_work}. The problem and relevant concepts are formalized in section~\ref{sec:problem}. The overall approach is described in section~\ref{sec:approach} and we analyze its  computational complexity in section~\ref{sec:compute}. Finally, in sections \ref{sec:exp_setup} to \ref{sec:evaluation} we quantitatively evaluate the approach on real-world data in terms of accuracy and used compute, as well as showings its robustness on a custom data set with some rare objects such as in Fig.~\ref{fig:teaser}. We discuss our conclusions in \ref{sec:conclusion}. 
%\newpage

% ===============================================================================================
\vspace{-12pt}
\section{Related Work} 
\label{sec:rel_work}
We here categorize related work based on their primary method of segmenting moving objects in point clouds: appearance-based segmentation of each point cloud, scan-to-scan change detection against the previous points, or map-based change detection by comparing the point cloud against a map. 
Several works make use of more than one method. 
Contrary to our focus on autonomous robots in unstructured cluttered environments, most recent work focus on  autonomous driving settings.

\rev{The spatial appearance of dynamic objects can be used for detection from only one point cloud.
This is} object- or environment-specific, either via a model \cite{petrovskaya2009model}, or via training on labeled examples in advance. 
A subtle but common appearance-based assumption is filtering out the ground \cite{petrovskaya2009model,Arora2021stachniss}. 
However, while streets and floors are mostly flat, this does not hold in many robotic applications. 
Many recent works focus on learning semantic segmentation of point clouds \cite{chen2021moving, sun2022efficient,pfreundschuh2021dynamic}, most focusing on autonomous driving and using SemanticKITTI \cite{behley2019semantickitti}.
However, a major limitation of appearance-based methods is that robots oftentimes have to operate in open-set environments, where previously unknown objects may be encountered.
Instead, we develop an approach to detect diverse moving objects in complex scenes.

It is \rev{also} possible to segment moving objects from residuals in point cloud registration or scene flow. 
Such approaches are agnostic of object type, but they are typically computationally expensive and struggle with slow-moving objects or articulated objects with separately moving parts, like pedestrians. 
Dewan \etal \cite{dewan2016motion} sequentially estimate multiple rigid motion hypotheses using RANSAC on point cloud registration. 
This has also been combined with learning appearance in \cite{dewan2017deep}.  
\rev{State-of-the-art} learning-based approaches for autonomous driving also include scan-to-scan depth residuals to improve an appearance-based approach \cite{chen2021moving, sun2022efficient}. 
\rev{However, learning based on appearance} loses the object agnosticity of \rev{pure} scan-to-scan approaches. By contrast, \rev{4DMOS \cite{mersch2022receding} recently proposed to train entirely on a sequence of point clouds to make it less dependent on object appearance.}

Map-based approaches typically require an existing map for detection. 
If an environment is already fully mapped, background subtraction can be used to aid recovery of moving objects \cite{Gehrung2017}.
Recent map-based approaches \cite{pfreundschuh2021dynamic,chen2022automatic} focus on generating labels for appearance-based classifiers offline. 
However, these are far too slow to run online and exploit that they have both past and future data on the scene.

Similarly, map cleaning approaches attempt to remove dynamic objects offline to compute accurate static maps of the environment \cite{lim2021erasor, schauer2018peopleremover, Arora2021stachniss}. 
While related, maps are typically dense enough even with considerable spurious removals, which means their metrics focus on static points and the approaches not directly comparable. 
These can make use of very approximate free-space calculations \cite{yoon2019mapless,pomerleau2014long}, and  recently, online cleaning of submaps was shown \cite{fan2022dynamicfilter}, but it only needed to clean when a submap of multiple point clouds was integrated rather than at sensor rates.

Most closely to ours, \cite{modayil2008initial} build an occupancy grid online, detecting points in free space as moving and cluster them into objects that are tracked through time. 
However it only uses 2D range finder scans which may not work well in 3D. 
In \cite{azim2012detection} an octree structure is used online to detect moving objects on streets but they note their approach  works poorly on pedestrians and provide no quantitative results.

%===============================================================================================
\vspace{-8pt}
\section{Problem Statement}
\label{sec:problem}

This work addresses the problem of real-time moving object detection.
Given an input point cloud $P_\mathcal{S}^{(t)}$ in sensor frame $\mathcal{S}$ at time step $t$, $P_\mathcal{S}^{(t)} = \{p_i\}, \ p_i \in \mathcal{P}$ and an estimated transform $\hat{T}_{\mathcal{W}\mathcal{S}}^{(t)}$ from sensor $\mathcal{S}$ to world frame $\mathcal{W}$. 
Without loss of generality, we use $\mathcal{P} =\mathbb{R}^3$ in this work. 
The goal is to segment all dynamic points $P_{dyn} \subseteq P_\mathcal{S}^{(t)}$, i.e.
\begin{align}
    & \max_{P_{dyn}} \frac{P_{dyn} \cap P^\star_{dyn}}{P_{dyn} \cup P^\star_{dyn}}, \text{ where}\label{eq:prob_goal} \\ 
    & P_{dyn} = F(P_\mathcal{S}^{(t)}, \dots, P_\mathcal{S}^{(0)}, \hat{T}_{\mathcal{W}\mathcal{S}}^{(t)}, \dots, \hat{T}_{\mathcal{W}\mathcal{S}}^{(0)}), \label{eq:prob_online} \\
    & P^\star_{dyn} = \{ p_i \in P^{(t)} \ |\ \sum_{t' < t}\lVert \Omega_\mathcal{W}^{(t')}(p_i) - \Omega_\mathcal{W}^{(t)}(p_i) \rVert > 0 \}. \label{eq:prob_gt} 
   \vspace{-5pt}
\end{align}
Here $\Omega_\mathcal{W}^{(t)}(p)$ is the position of the surface that generated $p$ in $\mathcal{W}$ at time $t$.
Intuitively, this corresponds to maximizing the Intersection over Union (IoU) between detected and true dynamic points (\ref{eq:prob_goal}).
As the robot operates online, the motion detector $F$ is a function of at most all previous inputs (\ref{eq:prob_online}).
We consider points as dynamic if the surface that has generated them has moved with respect to the world frame (\ref{eq:prob_gt}).
It is worth pointing out that we consider the general case where sensor readings may be imperfect, i.e. $\lVert T_{\mathcal{W}\mathcal{S}}^{\star(t)} p_i - \Omega_\mathcal{W}^{(t)}(p_i) \rVert \geq 0$, and the state estimates may drift during online robot operation, i.e. $\lVert \hat{T}_{\mathcal{W}\mathcal{S}}^{(t)} p_i - T_{\mathcal{W}\mathcal{S}}^{\star(t)} p_i \rVert \geq 0$ and increasing with time.
Lastly, the time to evaluate $F(\cdot)$ should not exceed the time between frames $t$ and $t+1$.

% ===============================================================================================
\vspace{-10pt}
\section{Approach}
\label{sec:approach}

\begin{figure}
    \centering
    \includegraphics[width=\linewidth]{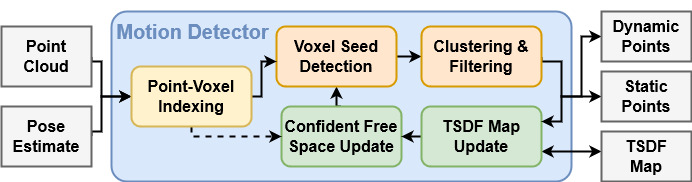}
    \caption{System overview. We first pre-process (yellow, Sec.~\ref{sec:app_preprocess}) each point cloud for efficient computation. Dynamic points are then detected (orange, Sec.~\ref{sec:app_detect}) in our high-confidence map and clustered to include low-confidence points. Lastly, we update (green, Sec.~\ref{sec:app_free}) the volumetric and free space maps with the new data.}
    \label{fig:system}
    \vspace{-15pt}
\end{figure}

To achieve this goal, we exploit the motion cue that when a point falls into space that is known to be free, the point must have moved there and thus be dynamic. 
However, knowing which spaces are truly free during online robot operation is a challenging problem.
Primary limitations include that the sensor readings may be sparse, the measured points noisy, the state estimate drifting from imperfect localization, and the reconstructed map inaccurate as observations are limited.
If space is wrongly classified as free, it will continuously detect static parts as moving.
If the detection is too conservative, the robot may fail to detect dynamic objects in time and jeopardize safety. 
To overcome these limitations, the central idea of our approach is to incrementally estimate a reduced but high confidence free space area by modeling each of these limitations. 
These high confidence areas are then used to seed dynamic object clusters and disambiguate points in low-confidence areas. 
As autonomous robots oftentimes have to build a map for navigation, we leverage this fact to present a light-weight implementation of our approach with minimal compute overhead over Voxblox \cite{oleynikova2017voxblox}, a widely used mapping framework.
An overview of our pipeline is shown in Fig.~\ref{fig:system}, and each component is further detailed below.

% ------------------------------------------------------------------------------
\vspace{-8pt}
\subsection{Map Representation}
A central component in mapping-based motion detection is the map representation.
In this work, we choose to adapt and extend Voxblox \cite{oleynikova2017voxblox}, as this has two major advantages:
First, the map required for safe navigation \cite{oleynikova2017voxblox} can directly be used for online motion detection, thus incurring little additional computation costs.
Second, we make extensive use of its hierarchical spatial hash-block structure for efficient real-time computation.
Nonetheless, other map representations are admissible to our approach.
This section briefly summarizes the relevant parts and presents our extensions of the map.

To incrementally map potentially unbounded environments, physical space is indexed as a grid of blocks $B = \{b_j\} \in \mathcal{B}$. 
Blocks touched by a sensor ray are dynamically allocated.
Each block contains a dense grid of $N_v$ voxels $v_k \in \mathcal{V}$, i.e. $b_j = \{v_0, \dots, v_{N_v}\}$. 
We use $N_v=16^3 = 4096$ and a voxel size $\nu=\SI{0.2}{m}$.
The map $M^{(t)}$ is the set of all allocated voxels $M^{(t)} = \rev{\cup_{j \in B^{(t)}}b_j^{(t)} = \{v_k^{(t)}\}}$.
To represent geometry, a Truncated Signed Distance Field (TSDF) as originally proposed by Curless and Levoy \cite{curless1996volumetric} 
is employed.
Each voxel $v^{(t)}$ at time $t$ has an associated signed distance $d$ and weight $w$.
This allows incremental updates with new measurements $d_{new}$ to filter out sensing errors:
\begin{align}
       & d^{(t+1)} = \frac{d^{(t)} \times w^{(t)} + d_{new} \times w_{new}}{w^{(t)} + w_{new}}, \label{eq:tsdf_d} \\
      & w^{(t+1)} = w^{(t)} + w_{new}.  \label{eq:tsdf_w} 
\end{align}
The weight $w_{new}$ represents the confidence of the current measurement.
As we evaluate our method on LiDAR data in this work, we use $w_{new}=1$ as they typically have constant measurement error \cite{os0}. 
This could easily be extended to e.g. inverse quadratic weights for depth cameras \cite{nguyen2012modeling}. 

However, the fact that TSDF fusion marginalizes out all temporal information makes it hard to employ it directly for temporal dynamics and motion detection.
We therefore augment the map with three compact attributes to represent the most relevant temporal aspects.
We summarize the temporal history of a voxel by the the last frame it was considered occupied $t_o \in \mathbb{N}$, and the duration of that occupancy in frames $t_d \in \mathbb{N}$.
As our method is based on reliable free-space estimation, each voxel has an attribute $f \in \{0,1\}$ denoting whether it is considered \emph{free} with high confidence. 
In summary, the TSDF map augmented with dynamic state is defined by the 5-tuple:
\begin{align}
    & v^{(t)} = \{ d^{(t)}, w^{(t)}, t_o^{(t)}, t_d^{(t)}, f^{(t)} \}. 
\end{align}
This formulation still allows efficient Markovian updates of the map, depending only on the current map $M^{(t)}$ and the incoming point cloud $P_\mathcal{W}^{(t)} = \hat{T}_{\mathcal{W}\mathcal{S}}^{(t)} P_\mathcal{S}^{(t)}$, and thus guaranteeing bounded memory usage that only scales with space:
\begin{align}
    & M^{(t+1)} = f(M^{(t)}, P_\mathcal{W}^{(t)}).
\end{align}

% ------------------------------------------------------------------------------
\vspace{-15pt}
\subsection{Integrating New Point Clouds}
\label{sec:app_preprocess}
In order to process an input point cloud $P^{(t)} = \{p_i\}, p_i \in \mathcal{P} = \mathbb{R}^3$ efficiently, we first build a mapping $I: \mathcal{P} \mapsto \mathcal{B}$, mapping the index of each point $p_i$ to its corresponding block index $b_j$. 
For each block, another mapping $J: \mathcal{P} \times \mathcal{B} \mapsto \mathcal{V}$ relates $p_i$ to its corresponding voxel index $v_k$ in $b_j$.
Naturally, the mapping $K: \mathcal{P} \mapsto \mathcal{V}$ relating points $p_i$ to voxels $v_k$ can be computed as $K = I \circ J$.
These map indexes are then used to efficiently detect dynamic points as described in Sec.~\ref{sec:app_detect} and update our map estimate as described in Sec.~\ref{sec:app_free}.

% ------------------------------------------------------------------------------
\vspace{-10pt}
\subsection{High-confidence Free-space Estimation}
\label{sec:app_free}
To enable robust motion detection with the limited information available during online operation, the main idea of our approach is to \rev{use} the high-confidence \emph{free} label $f$ to seed the detection of complete dynamic objects \rev{also in low-confidence areas}.
The high confidence estimate is achieved by modeling the effects of sparse sensor readings, noisy measured points, drifting state estimate, and inaccurate map reconstructions when observations are limited.

\textbf{Occupancy Estimation:}
To make use of both the temporally smoothed information in the map $M^{(t)}$ and the most recent information in $P^{(t)}$, we consider a voxel $v$ occupied at $t$ if its TSDF distance $d^{(t)}$ is below a conservative threshold $\tau_d=1.5\nu$ or if a current point $p_i$ falls into it:
\begin{equation}
    t_o^{(t+1)} = \begin{cases}
    t &\text{if } d^{(t)} < \tau_d \vee \exists p_i \in P^{(t)}: K(p_i) = v \\
    t_o^{(t)}& \text{otherwise.}
    \end{cases}
    \label{eq:cues}
    \vspace{-1pt}
\end{equation}

\textbf{Sensor Sparsity:}
However, sensor data may be sparse at long distances and not every voxel may be intersected by a ray, as e.g. even for a 128-beam Ouster OS0 sensor with 90$^\circ$ vertical field of view, the vertical point spacing at $\SI{20}{m}$ is $\sim\SI{0.25}{m} > \nu= \SI{0.2}{m}$.
To account for this possible sparsity, we allow for a sparsity compensation duration of $\tau_s=2$ frames before consecutive occupancy is terminated:
\begin{equation}
    t_d^{(t+1)} = \begin{cases}
    t_d^{(t)}+1 &\text{if } t_o^{(t)} \geq t - \tau_s \\
    0 & \text{otherwise.}
    \end{cases}
    \vspace{-1pt}
\end{equation}

\textbf{Temporal and Spatial Confidence:}
Given this occupancy information, we can now estimate the \emph{free} voxels.
However, as voxels wrongly labeled as free will produce many false positives and are hard to recover from, we employ a both spatially and temporally conservative estimate.
For temporal certainty to avoid errors arising from noise, we require a \emph{temporal window} $\tau_w$ to pass before a voxel can be considered free. 
As this parameter is hard to deduce from physical properties, we choose without further tuning $\tau_w=5$ as we assume 5 frames to be unlikely to be all noisy and $\SI{0.5}{s}$ is still an acceptable duration before objects moving into newly mapped space can be detected.
For spatial certainty, instead of checking individual voxels, we require the developed criterion to hold for the voxel and all its neighbors.
Lastly, as it is known that TSDFs can be inaccurate at the boundary of unknown space \cite{pfreundschuh2021dynamic}, since this space might be occupied, we also require all neighbor voxels to be observed:
 \vspace{-2pt}
\begin{align}
    & \hat{f} = \mathbb{I}\left(t_o^{(t)}(v') < t - \tau_w \wedge w^{(t)}(v') > 0, \ \forall v' \in \mathcal{N}(v) \right) \label{eq:free} \\
    & f^{(t+1)} = \max(f^{(t)}, \hat{f}) \label{eq:free_update}
 \vspace{-2pt}
\end{align}
where $\mathcal{N}(v)$ are the neighbors of voxel $v$, $t_o^{(t)}(v')$ and $w^{(t)}(v')$ are their weight and last time they were occupied, and $\mathbb{I}(\cdot)$ is the indicator function. 
Note that in (\ref{eq:free_update}) once a voxel has been observed to be free it stays free.
On one hand, this is important as free voxels turning occupied is the major motion cue exploited in this work.
On the other hand, this is problematic when the robot state estimate is not perfect.

\textbf{State Estimation Drift:}
As drift may occur over time, large parts of the static observations now fall into previously free space and are wrongly detected.
A central challenge in compensating drift is that both drift and dynamic objects are perceived as motion within the sensor frame.
However, a distinguishing feature is the velocity at which things move, as drift is typically characterized by slowly accumulating tracking errors.
To compensate for this, we allow voxels to turn occupied when the observed motion is slow and the voxels are located in the low-confidence free-space area, thus modifying (\ref{eq:free_update}) to:
\vspace{-5pt}
\begin{equation}
     f^{(t+1)} = \begin{cases}
     0 & \text{if } t_d^{(t)} > \tau_r \\
         \max(f^{(t)}, \hat{f}) & \text{otherwise.}
     \end{cases}
\label{eq:free_update_2} 
\end{equation}
To account for (\ref{eq:free}), once a voxel $v$ turns occupied, we also set $f(v')=0, \ \forall v' \in \mathcal{N}(v)$.
Considering the number of frames it takes a static point to drift through a voxel, the optimal reset duration $\tau_r$ can be computed based on an estimate of the maximum expected drift rate $\hat{r}_{max}$ and the sensor frame rate $h$ as:
    \vspace{-5pt}
\begin{equation}
    \tau_r =\nu  h {\hat{r}_{max}}^{-1}
    \label{eq:drift_param}
    \vspace{-2pt}
\end{equation}
As drifting static points always move from the occupied areas into the low-confidence areas which are then slowly turning occupied, they will not be detected as dynamic as long as $r^{\star(t)}<\hat{r}_{max}$.
However, consequently, objects moving slower than $\hat{r}_{max}$ may also not be detected as dynamic. 
We study these effects in detail in Sec.~\ref{sec:res_drift}.

% ------------------------------------------------------------------------------
    \vspace{-5pt}
\subsection{Detecting Dynamic Points}
\label{sec:app_detect}
Dynamic points are detected following the free-space motion cue.
In particular, if a previously free voxel is occupied, this means these points must have moved there and are thus dynamic. 
We leverage this principle to efficiently detect dynamic points $P_{dyn}$ in the map space by detecting their corresponding voxels $V_{dyn}$.
However, as $f(v)$ is a conservative estimate, we first compensate for the constraints introduced in (\ref{eq:free}).
Since we detect dynamic points in every frame, the temporal constraint does not need correction. 
The spatial constraint can efficiently be rectified by inverting the neighborhood constraint:
\begin{equation}
    V_{dyn} = \{ v_k \in K(P^{(t)}) \ |\ \exists v' \in \mathcal{N}(v_k) : f(v') = 1 \} 
        \vspace{-2pt}
\end{equation}
Conceptually, this corresponds to using the high confidence detection ($f=1$) to seed dynamic objects, and then grow these initial seeds to also include adjacent points in low-confidence areas.
We extract object clusters by grouping all dynamic voxels $V_{dyn}$ into connected components $C =\{c_l\}$.
To avoid detecting individual noisy points, we filter out clusters where $|c_l| < \tau_c=20$. 
The final set of dynamic points is thus given as:
\begin{equation}
    P_{dyn} = \{ p_i \in P^{(t)} \ |\ K(p_i) \in C \}
        \vspace{-2pt}
\end{equation}
It is worth noting that we focus on frame-wise detection and thus no additional temporal tracking and filtering is applied. 
Nonetheless, such tracking could readily be combined with our method to further improve performance in future work.

Lastly, to account for dynamics also in the geometry representation of the map, we set $w(v)=0, \ \forall v \in C$ during the next update.
This leads (\ref{eq:tsdf_d}) and (\ref{eq:tsdf_w}) to overwrite rather than average voxels known to be dynamic, thus representing the most up-to-date information in dynamic areas while filtering out noise in static parts.

% ------------------------------------------------------------------------------
    \vspace{-10pt}
\section{Computational Complexity}
\label{sec:compute}
Real-time detection of dynamic objects during online robot operation is essential for the safety of the robot and its environment. 
To this end, we analyze the computational complexity as well as paralellizability of our approach.
The presented method can be split into 4 operations: pre-processing $F_{pre}$, clustering $F_{clust}$, free space estimation $F_{free}$, and TSDF integration $F_{tsdf}$.
During pre-processing, we build the mappings $I, J, K$ by looking up the block and voxel index of each point in $P^{(t)}$.
Due to the hash implementation, both operations are $\mathcal{O}(1)$:
\begin{equation}
    \mathcal{O}(F_{pre}) = \mathcal{O}(|P^{(t)}|) = \mathcal{O}(1)
    \label{eq:comp_pre}
        \vspace{-2pt}
\end{equation}
The second equality holds as the number of points per scan $|P^{(t)}|$ is typically constant for a given sensor.
Note that during this lookup we can simultaneously detect all cluster seeds $V_{seed}^{(t)} = \{ v \in K(P^{(t)}) \ |\ f(v) = 1 \}$.
Since this is a non-modifying access it is completely data-parallel, although we run it in a single thread in our implementation. 
As we perform clustering directly in map space (Sec.~\ref{sec:app_detect}) and neighbor voxels can be computed in $\mathcal{O}(1)$, this is implemented as a wavefront search with worst-time complexity of:
\begin{equation}
    \mathcal{O}(F_{clust}) = \mathcal{O}(|V_{seed}^{(t)}|)
        \vspace{-2pt}
\end{equation}
Note that while this operation is not very optimized, $|V_{seed}|$ is typically low in practice and $F_{clust}$ only a minor contribution to the overall computation.
To update the map, we perform a projective TSDF update for all blocks touched by the current measurement, denoted $\bar{B}^{(t)}$.
Afterwards, each block is updated in parallel in constant time:
\begin{equation}
    \mathcal{O}(F_{tsdf}) = \mathcal{O}(F_{free}) = \mathcal{O}(|\bar{B}^{(t)}|)
    \label{eq:comp_map}
        \vspace{-2pt}
\end{equation}
Lastly, in order to guarantee real-time performance, the computational cost of the complete algorithm combining equations (\ref{eq:comp_pre})-(\ref{eq:comp_map}) can be bounded.
We note that in the worst case $|V_{seed}^{(t)}| \leq |P^{(t)}|$. 
As $|P|$ is a sensor parameter, $F_{pre}$ and $F_{clust}$ generally consume a fixed but comparably low computation amount.
On the other hand, $F_{tsdf}$ and $F_{free}$ scale $\propto \mathcal{O}(|\bar{B}^{(t)}|)$, where $|\bar{B}^{(t)}|$ due to its volumetric nature scales $\propto (d_{int} / \nu)^3$ for an integration distance $d_{int}$.
It is also standard to use a maximum integration distance $d_{int}$ for real-time application of volumetric mapping in large areas, such that $|\bar{B}^{(t)}|$ is bounded $<\bar{B}_{max}$ and a constant maximum computation time can be guaranteed. 
We discuss this effect in detail in Sec.~\ref{sec:res_compute}.
\vspace{-10pt}
% ===============================================================================================

\section{Experimental Setup}
\vspace{-2pt}
\label{sec:exp_setup}
\textbf{Datasets:}
We quantitatively evaluate our approach on the DOALS \cite{pfreundschuh2021dynamic} dataset, featuring 8 sequences in 4 environments captured with a high-range, high-resolution OS1 64 LiDAR at $h=\SI{10}{Hz}$.
However, since the DOALS data primarily explores open and flat environments with pedestrians as moving objects, we further present qualitative results on several newly recorded sequences of challenging environments and varying moving objects.
To this end, we collect data with an OS0 128 high-range, high-resolution LiDAR at $h=\SI{10}{Hz}$ \cite{os0}. 
The data is processed with Fast-lio2 \cite{xu2022fast} to achieve realistic state estimates during robot operation. 
We publicly release these additional sequences with our method.

\textbf{Metrics:}
In each sequence of DOALS, 10 frames were manually annotated. 
For each of these frames, we compute the IoU (\ref{eq:prob_goal}) between all points in $P^{(t)}$ that were labeled as dynamic in the data set $P^\star_{dyn}$, and the points $P_{dyn}$ detected for this frame, and report the mean over all 10 frames.

\textbf{Hardware:}
To ensure real-time applicability, we run all experiments on a NUC with laptop-grade AMD-4800U CPU also used in some of our aerial and ground robots.
% ===============================================================================================
\vspace{-12pt}
\section{Evaluation}
\label{sec:evaluation}

\begin{table}
\vspace{-3pt}
\adjustbox{max width= \linewidth}{
\setlength{\tabcolsep}{3pt}
\centering
    \begin{tabular}{lccccc}
    Method $\setminus$ Dataset & Station & Shopville & HG & Niederdorf & All \\  
    \midrule
    Occupancy \cite{pfreundschuh2021dynamic} {\footnotesize(Offline)} & 91 & 85 & 88 & 87 & 88\\
    \midrule
    \rev{DOALS-3DMiniNet} \cite{alonso20203d, pfreundschuh2021dynamic} & 84 & 82 & 82 & 80 & 82\\
    \rev{4DMOS \cite{mersch2022receding}} & \rev{38.8} & \rev{50.6} & \rev{71.1} & \rev{40.2} & \rev{50.2}\\
    \rev{LMNet \cite{chen2021moving} {\footnotesize(Original)}} & \rev{6.0} & \rev{7.5} & \rev{4.6} & \rev{3.0} & \rev{5.2}\\
    \rev{LMNet \cite{chen2021moving} {\footnotesize(Refit)}} & \rev{19.9} & \rev{18.9} & \rev{27.4} & \rev{40.1} & \rev{26.6}\\
    \rev{MotionSeg3D \cite{sun2022efficient}} & \rev{\ding{55}} & \rev{\ding{55}} & \rev{\ding{55}} & \rev{\ding{55}} & \rev{\ding{55}}\\
    Ours & \textbf{86.2} & \textbf{83.2} & \textbf{84.1} & \textbf{81.6} & \textbf{83.8} \\
   % Ours {\footnotesize(20m $\inf$)}   & 81.6 & 77.8 & 69.2 & 82.4 & 77.7 \\
    \midrule
    LC Free Space \cite{modayil2008initial} {\footnotesize(20m)} & 48.7 & 31.9 & 24.7 & 17.7 & 30.7 \\
    Ours {\footnotesize(20m)} & \textbf{87.3} & \textbf{87.8} & \textbf{86.0} & \textbf{83.1} & \textbf{86.0} \\
    \bottomrule
    \end{tabular}
    }
    \caption{Dynamic point detection IoU [\%] on the DOALS dataset. }
    \label{tab:baselines}
    \vspace{-15pt}
\end{table}

% NOTE: without the grid it's hard to tell scenes apart
\begin{figure*}
\centering
\setlength{\tabcolsep}{0pt}
\begin{tabular}{|c|c|c|}
\hline
\includegraphics[width = 0.33\linewidth]{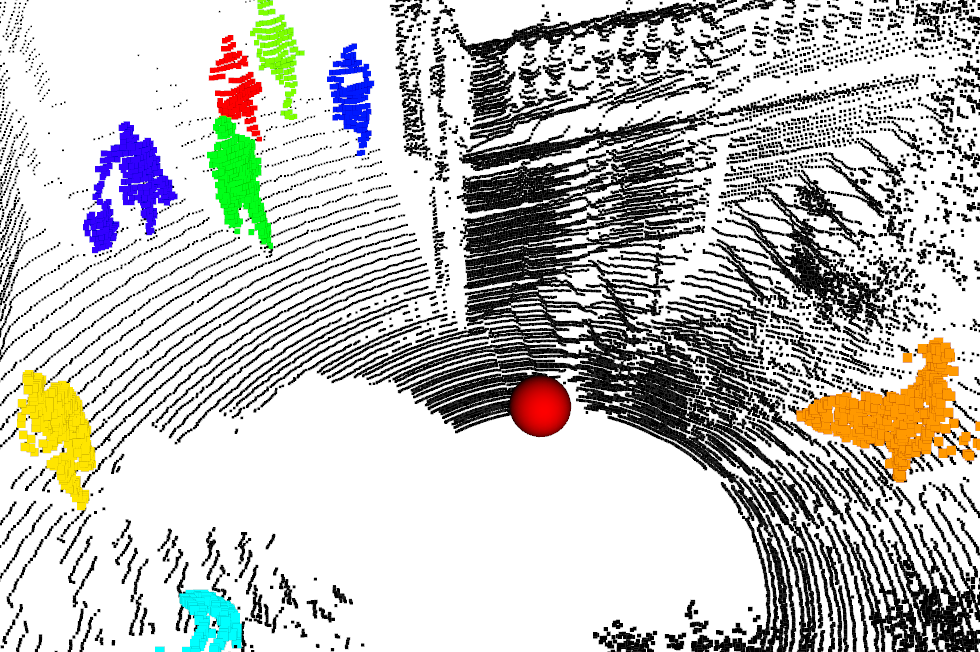} &
\includegraphics[width = 0.33\linewidth]{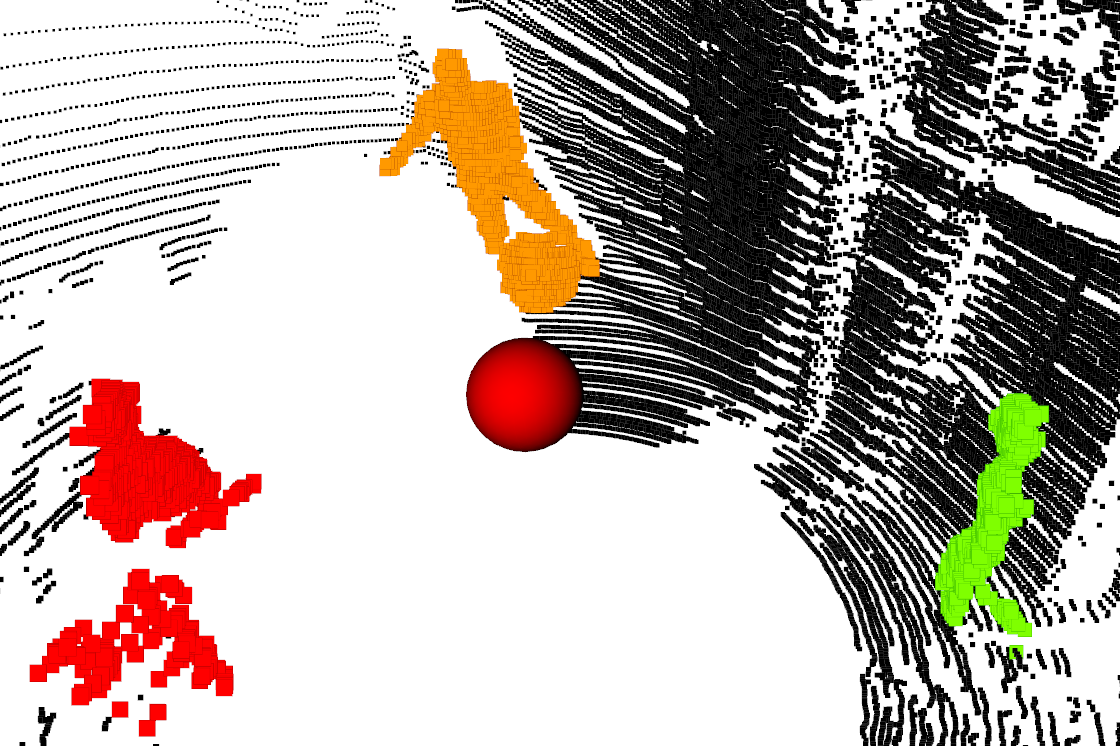} &
\includegraphics[width = 0.33\linewidth]{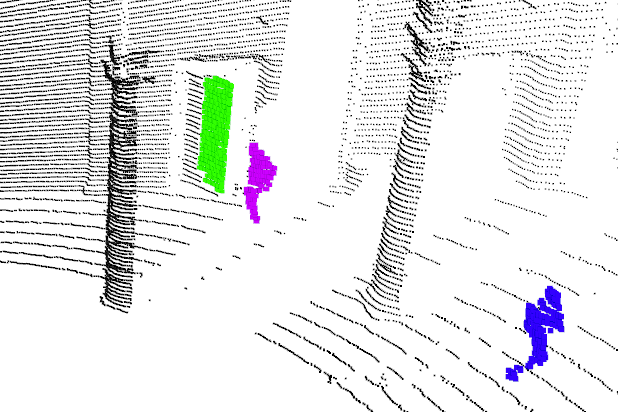}\\[-4pt]
\hline
\includegraphics[width = 0.33\linewidth]{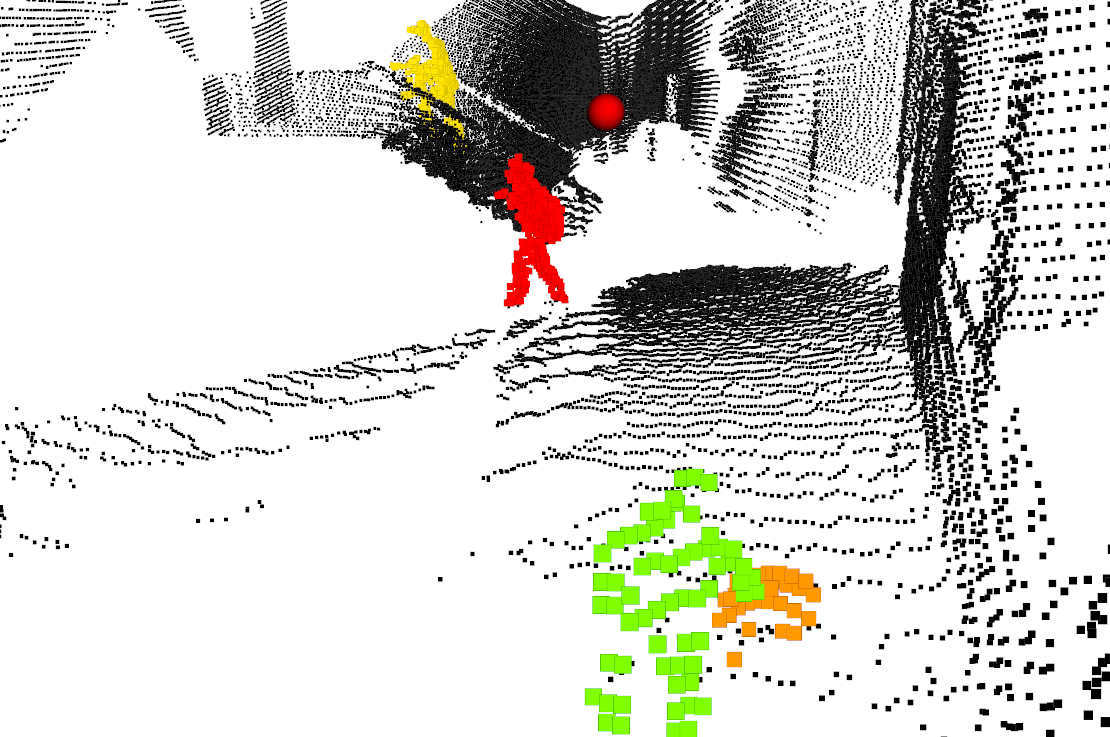}&
\includegraphics[width = 0.33\linewidth]{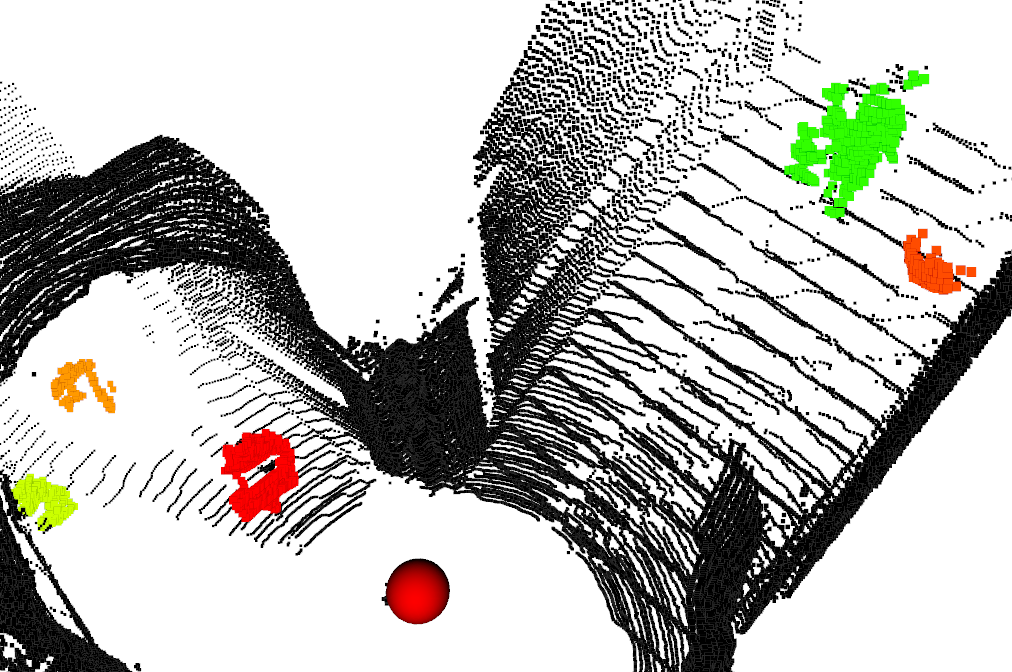}&
\includegraphics[width = 0.33\linewidth]{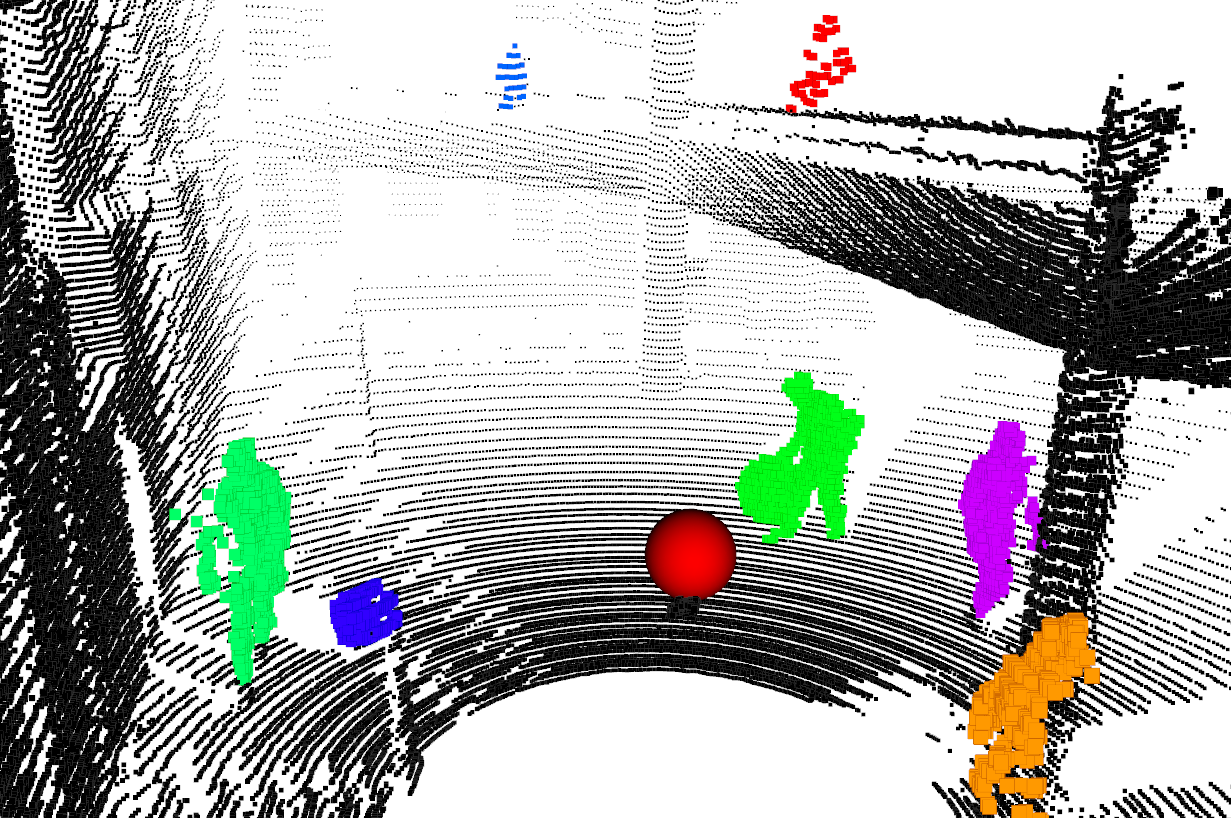}\\[-4pt]
\hline
\end{tabular}%
\caption{Qualitative detection results of our method. The sensor is shown as \quali{ff0000}{red sphere}. Top row: Challenging object classes such as a \quali{ff9900}{person with surfboard}, \quali{3200ff}{person with rolling case}, or \quali{00ffff}{rolling ball} (left), \quali{ff9900}{football player} (center), or \quali{33ff00}{door} after being pushed open by a \quali{cc00ff}{person} (right). Bottom row: Challenging environments where the sensor and pedestrians are on \textbf{stairs outdoor} (left), in a \textbf{narrow staircase} (center), or distributed across \textbf{multiple stories}. Sensor points are colored by cluster if detected dynamic and black if static.}
\label{fig:qualitative}
\vspace{-19pt}
\end{figure*}

\vspace{-5pt}
\subsection{Dynamic Point Detection Performance}
First, we evaluate the detection performance of our method and several baselines on the DOALS data in Tab.~\ref{tab:baselines}.
We compare against the recent \emph{Occupancy}-based approach of \cite{pfreundschuh2021dynamic}. 
Note that Occupancy is an \emph{offline} approach with infinite compute time and complete \rev{knowledge}, i.e. \rev{compared to online methods it has access to all past \emph{and future} scans}, and is thus used as an upper performance limit.
We further compare against \rev{recent learning-based approaches DOALS-3DMiniNet \cite{alonso20203d, pfreundschuh2021dynamic}, 4DMOS \cite{mersch2022receding}, LMNet \cite{chen2021moving}, and MotionSeg3D \cite{sun2022efficient}. These are evaluated on the \emph{full range}, reaching up to $\SI{172.7}{m}$.}
However, in our main application of robot autonomy, \rev{local context is frequently sufficient}. 
We therefore also evaluate with a standard maximum range of $d_{max}=\SI{20}{m}$, \rev{and compare to another mapping-based approach that directly queries the \emph{Low-confidence (LC) Free Space} \cite{modayil2008initial}.}

We observe in Tab.~\ref{tab:baselines} that our method shows strong performance even on the full-range data, outperforming the \rev{learning-based methods} and approaching the hindsight offline method. 
We note that this is a particularly challenging domain for several reasons. 
\rev{First, DOALS-3DMiniNet has been trained on the other environments of the DOALS dataset and learnt to identify pedestrians, which make up the vast majority of dynamic objects. 
It achieves best performance among the learning-based methods.
\cite{mersch2022receding,chen2021moving,sun2022efficient} are pre-trained on KITTI \cite{behley2019semantickitti}, as no labeled training data for DOALS is available. 
While 4DMOS, which similar to ours emphasizes spatial motion cues, transfers to some degree to the new dataset, the more appearance-based methods LMNet (Original) and MotionSeg3D achieve low performance. 
To reduce this domain gap, we further refit the model statistics of LMNet and MotonSeg3D on the DOALS data (labeled \emph{Refit}). 
While this improves performance, it is important to point out that this is model fitting on the evaluation data, and performance is still substantially lower than ours.
We did not achieve meaningful results for MotionSeg3D even with refitting.
This suggests that, while these methods show strong performance in their original domain, they can be hard to generalize to out-of-domain data sets without significant retraining.
}
In contast, our approach \rev{is completely prior-free and} object class agnostic. 
\rev{This is also reflected in the lower variance across sequences.}
Second, the density of the LiDAR scan and therewith also quality of the map notably decreases with higher distances.
This is reflected in the slightly improved performance if only $d_{max}=\SI{20}{m}$ \rev{are} considered, as detecting nearby dynamic objects is usually more relevant.
Nonetheless, our high IoU on full range highlights the applicability of our method to long-range detection scenarios.
Lastly, we observe that using the conventional free-space \rev{\cite{modayil2008initial}} results in significantly reduced performance\rev{, reinforcing }the importance of our high-confidence map.

% --------------------------------------------------------------------------------------------
\vspace{-6pt}
\subsection{Robustness to Object Class and Environment} 
To evaluate the capacity of our approach to detect diverse moving objects in challenging environments, we show qualitative detection results of our recorded sequences in Fig.~\ref{fig:qualitative}.
A large advantage of our proposed method is that, since it is purely geometric, it is completely agnostic to the class of moving object. 
This stands in stark contrast to most current learning-based approaches that are oftentimes pre-trained on specific objects such as pedestrians \cite{chen2021moving, sun2022efficient, pfreundschuh2021dynamic}, which may not generalize well to the open-set world encountered by robots.
We observe that our method correctly detects numerous pedestrians, also with unusual appearances, e.g. when carrying boxes, rolling cases, or even surfboards.
We further successfully detect radically different objects such as balls rolling around or swinging doors.
In addition, we observe that our method is applicable to various challenging and unstructured scenes, where assumptions about the environment such as the existence of a ground plane may not hold.
Notably, due our conservative estimate, also challenging geometry is correctly classified as static.

% -------------------------------------------------------------------------------------------
\vspace{-6pt}
\subsection{Robustness to Drift}
\label{sec:res_drift}

\begin{figure*}
    \centering
    \includegraphics[width=0.32\linewidth]{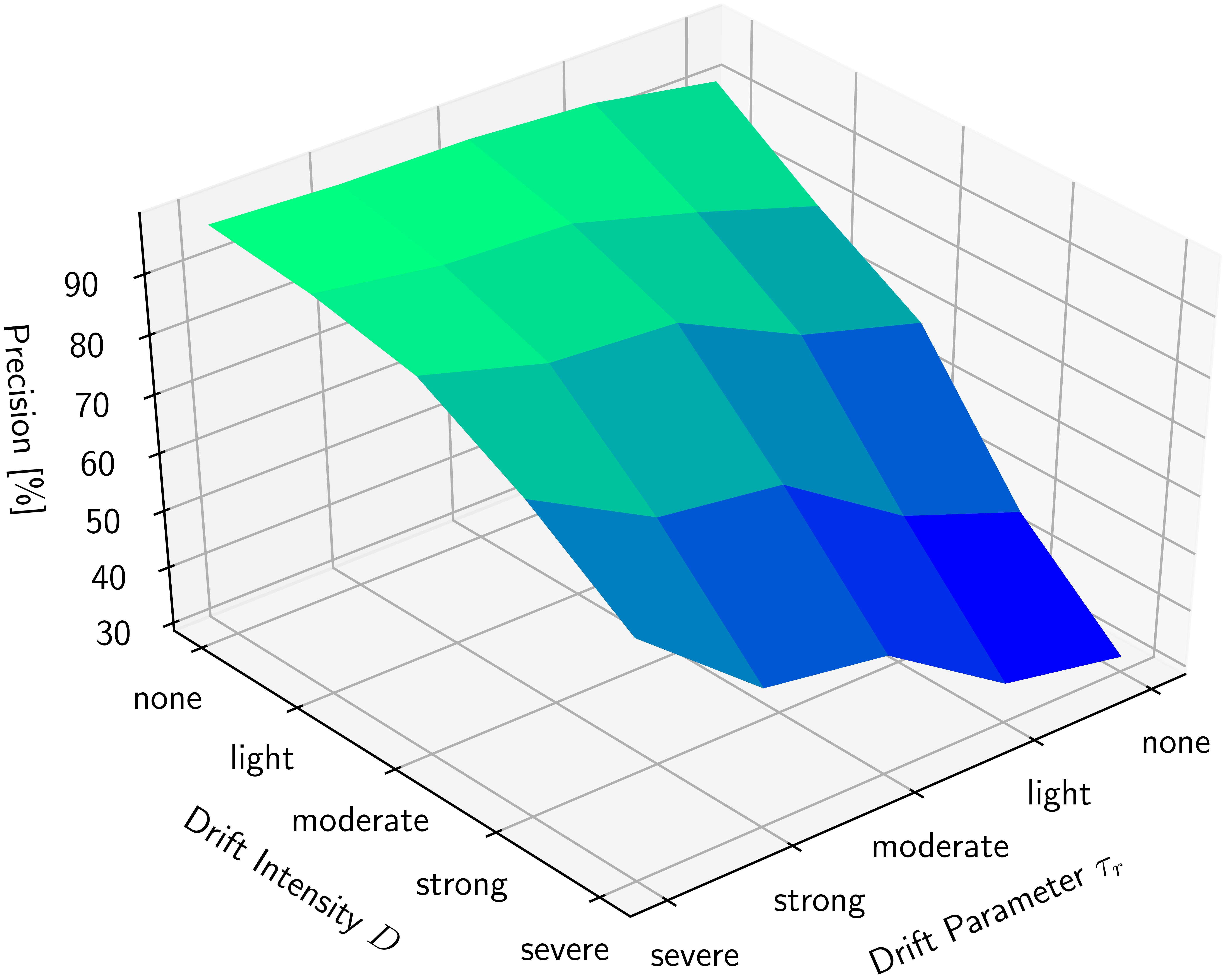}
    \includegraphics[width=0.32\linewidth]{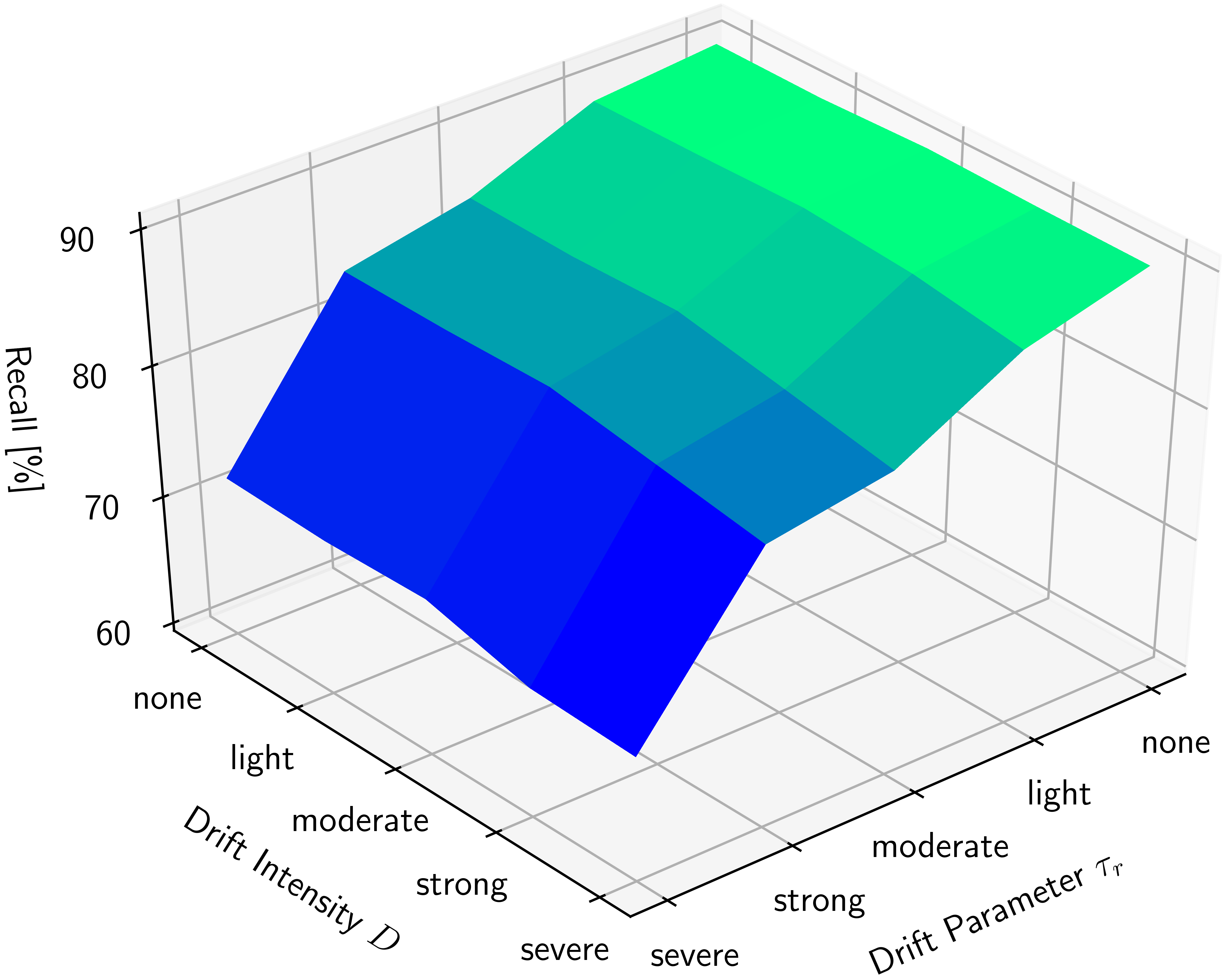}
    \includegraphics[width=0.32\linewidth]{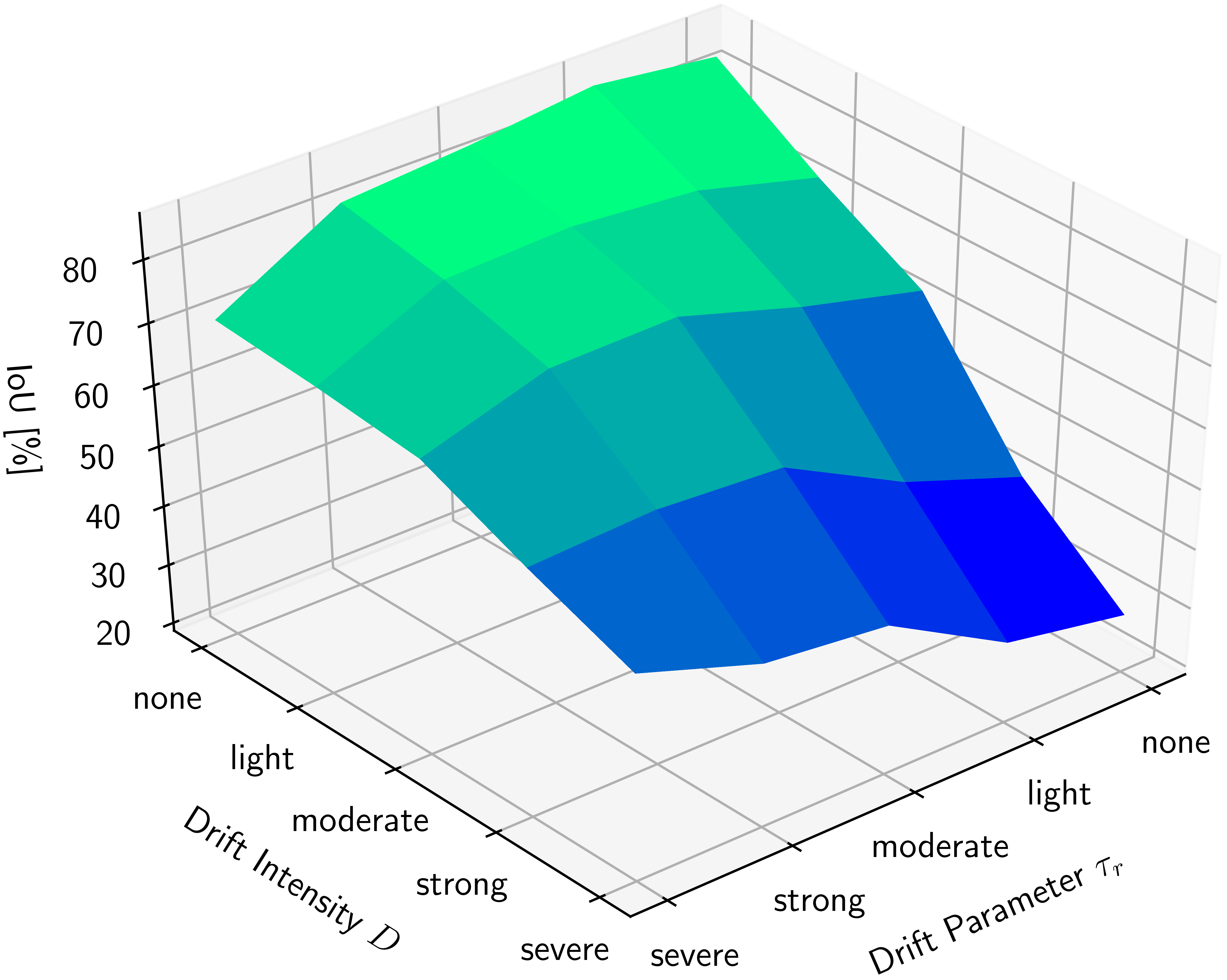}
    \caption{Precision (left), recall (center), and IoU (right) for increasing drift intensities and parameter choices.}
    \label{fig:drift}
    \vspace{-20pt}
\end{figure*}

To study the robustness of our method to imperfect state estimates and the importance of the drift parameter $\tau_r$, several experiments on the DOALS dataset were conducted.
We use the realistic drift simulator of \cite{schmid2021unified} to generate 3 random drift rollouts for each data sequence and each intensity $D \in \{\text{\emph{none, light, moderate, strong, severe}}\}$.
\rev{This results in true maximum drift rates $r_{max}^\star= \{1.26, 3.65, 4.54, 10.38\}\ \SI{}{cm/s}$, representing $\sim$1-10\% of the original motion.}
Note that %\emph{moderate} drift represents drift levels that can be expected by state of the art visual-inertial odometry systems \cite{schmid2021unified}, and that 
$D$ follows a $\log_2$-scale, i.e. severe drift is 4x stronger than moderate to \rev{limit-test our approach}.
To adequately evaluate the effect of $\tau_r$, we compute $\tau_r$ according to (\ref{eq:drift_param}) as $\tau_r = \{\infty, 150, 50, 40, 15\}$, respectively. 
%During robot deployment, the expected maximum rate $\hat{r}_{max}$ is instead used to set $\tau_r$.
Fig.~\ref{fig:drift} shows the precision, recall, and IoU for all combinations of $D$ and $\tau_r$.
We observe that, as expected, the precision increases for more conservative $\tau_r$, and also decreases for higher drift as many false positives are detected.
Although avoiding false positives is a challenging problem at high $D$, we observe that the effect can be mitigated by $\tau_r$, reducing the decrease in precision by up to a factor of $2.9\times$.
We find that the recall is independent of the drift intensity, highlighting that most truly moving points can still be detected in the presence of drift. 
Increasing values of $\tau_r$ result in lower recall, as some slowly moving objects are no longer correctly detected.
However, we note that this effect is much less pronounced than the decrease in precision, leading to a recall of still $72\%$ in the worst case.
Combining all effects in the final IoU, we observe a ridge of optimal performance when $\tau_r=D$, validating our modeling assumptions.
%We further note that the ridge is comparably flat in $\tau_r$, highlighting that the approach is not very sensitive to the exact choice of $\tau_r$.
\rev{This supports the adequacy of our map-based approach, showing strong performance with \emph{light} drift similar to that of recent LiDAR-Odometry \cite{xu2022fast}, and showing robustness at up to 3.5m displacement in our \emph{severe} experiments.}
For practical considerations, we note that setting $\tau_r$ correctly can improve performance by up to 88\%, with a maximum performance drop of 15\% when overestimating $\tau_r$, suggesting the use of a slightly higher $\tau_r$ when in doubt.
% Minor exception: for drift param none (param=$\infty$), the groundtruth data is not completely drift free and additional sensing noise, so slightly worse than p=light. We thus recommend always using a slight drift param.

%------------------------------------------------------------------------------------------------
\vspace{-6pt}
\subsection{Computation Cost Analysis}
\label{sec:res_compute}

\begin{figure}
    \centering
    \includegraphics[width=\linewidth]{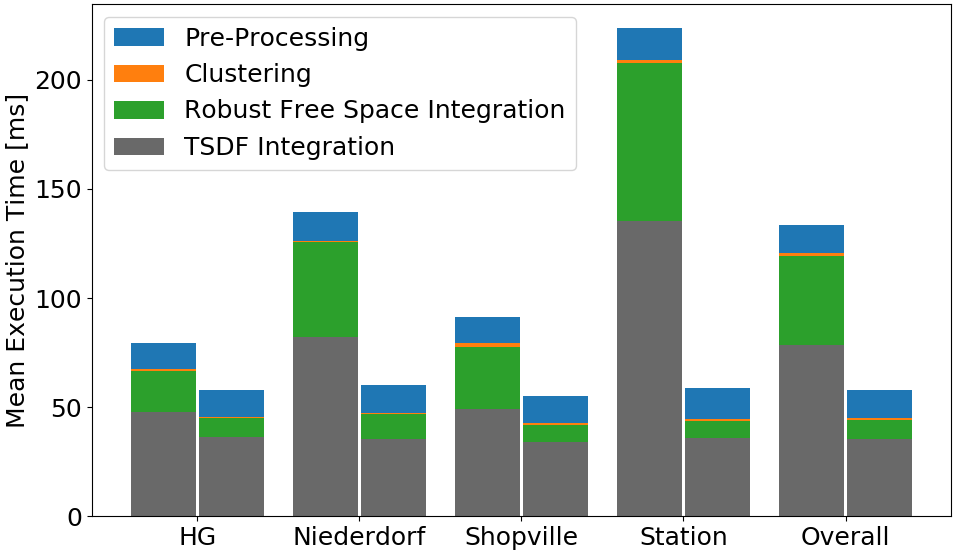}
    \vspace{-10pt}
    \caption{Average computation times on different datasets for \emph{infinite range} (left) and \emph{20m} range (right), split by algorithmic component.}
    \label{fig:compute}
\end{figure}

As  real-time computation on-board mobile robots is essential for safety, we empirically study the compute cost of our system in Fig.~\ref{fig:compute}, for $d_{int} = \infty$ (left column) and $d_{int}=\SI{20}{m}$ (right column).
We observe that in accordance with our analysis in Sec.~\ref{sec:compute}, the time for pre-processing is constant $13.0 \pm 0.8 $ms. 
Note that this could be further optimized as it is fully parallelizable.
While clustering time varies, the total compute consumption is negligible.
Lastly, we observe that both $f_{free}$ and $f_{tsdf}$ notably scale with the amount of blocks $|\bar{B}^{(t)}|$ updated. 
This is most prominent in large open spaces with $d_{int}=\infty$, such as in the Station environment.%, measured with a long-range, high-resolution sensor leads to longer computation times. 
We empirically find that $f_{tsdf} = 20.2 + 1.49 \times f_{free}$ is a virtually perfect fit, corroborating the linear scaling relationship.
Notably, by considering $d_{int} = \SI{20}{m}$ the overall run-time is reduced to a consistent $58.1 \pm 1.9 \SI{}{ms}$, leading to a frame rate of $17.2$ FPS for online robot operation. 
We note that regular TSDF integration takes up the majority of the computation time.
Thus, if a volumetric map is anyways needed for navigation, our approach results only in an added computation cost of 38.8\%, making it well suitable for lightweight integration when navigating dynamic environments.

% -----------------------------------------------------------------------------------------------

\subsection{Ablation Study}
\vspace{-2pt}

\begin{table}
\setlength{\tabcolsep}{3pt}
\adjustbox{max width= \linewidth}{
\centering
    \begin{tabular}{lcc|lcc}
    Method & IoU & $\Delta$ & Method & IoU & $\Delta$ \\  
    \midrule
    Ours & \textbf{86.0} & \textbf{0.0} & Ours & \textbf{86.0} & \textbf{0.0} \\
     {\footnotesize w/o} Occupancy Cue & 85.6 & -0.4 & {\footnotesize w/o} Sparsity Comp. $\tau_s$ & 83.3 & -2.7 \\
     {\footnotesize w/o} TSDF Cue & 23.6 & -62.4 & {\footnotesize w/o} Spatial Margin $\mathcal{N}$ & 38.1 & -47.9 \\
     {\footnotesize w/o} temporal window $\tau_w$ & 11.6 & -74.4 & {\footnotesize w/o} Cluster Filter $\tau_c$ & 85.3 & -0.7 \\
    \bottomrule
    \end{tabular}
    }
    \caption{Ablation study of the proposed components at 20m [\%].}
    \label{tab:ablation}
    \vspace{-12pt}
\end{table}

We empirically study the importance of the various effects modeled in Sec.~\ref{sec:approach}, presented as an ablation study in Tab.~\ref{tab:ablation}.
While each modeled effect contributes to the overall performance, we observe a clear split between aspects central to the method that lead to stark performance drops, and minor effects that only polish the final performance.
In terms of mapping cues (\ref{eq:cues}), we observe that fusion of measurements in the TSDF map is an essential component for robust detection, whereas the up-to-date occupancy cues make only a minor difference.
This may be explained by the fact that we update the TSDF to be up-to-date in dynamic regions, and by the comparably slow motions of pedestrians in the dataset which are already well captured in the TSDF.
We find that being conservative in classifying space as free, both temporally ($\tau_w$) and spatially ($\mathcal{N}$) is crucial, as getting wrong labels once lead to numerous false positives in the future.
For this reason, we note that erasing the spatial or temporal confidence leads to even worse performance than directly querying the TSDF map (Tab.~\ref{tab:baselines}), whereas combining both almost triples performance.
Lastly, we find that modeling sensing sparsity ($\tau_s$) and outlier points ($\tau_c$) does improve performance, but not by a large margin compared to the other components.
This corroborates our assumptions that the volumetric map can be a powerful tool to detect moving objects, if free space is modeled with high confidence.

% -----------------------------------------------------------------------------------------------
\subsection{Limitations}
\vspace{-2pt}
\begin{figure}
    \centering
    \includegraphics[height=90pt]{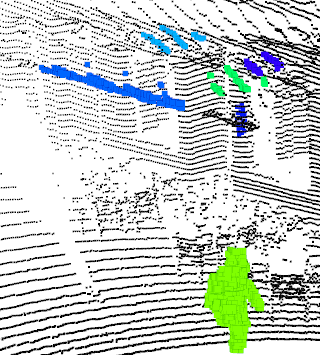}
    \includegraphics[height=90pt]{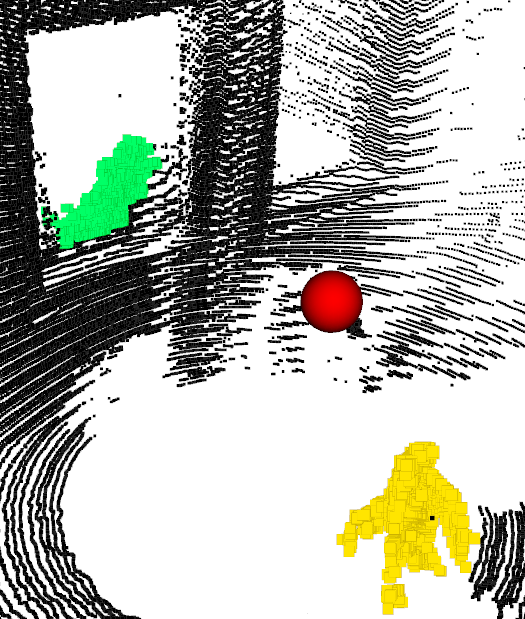}
    \includegraphics[height=90pt]{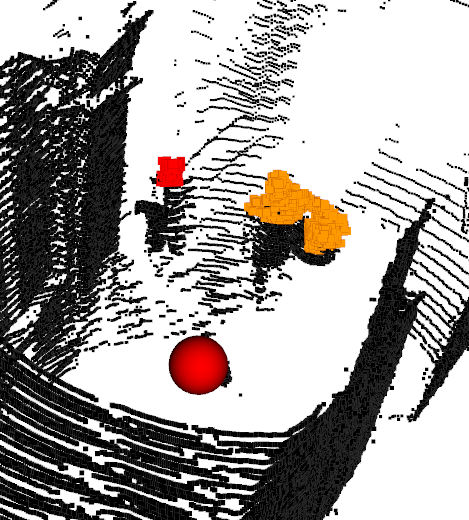}
    \caption{Limitations can include sparse thin surfaces (left), reflective surfaces (center) and strong occlusions during mapping (right).}
    \label{fig:limitation}
    \vspace{-16pt}
\end{figure}

Occasional failures are shown in Fig.~\ref{fig:limitation}. 
Extremely thin and sparsely measured objects such as a shade are hard to represent in the voxel-based map.
This issue could be alleviated by e.g. using neural map representations \cite{lionar2021neuralblox} that can achieve high resolution to model thin objects.
Reflective surfaces such as windows are known to be problematic for LiDAR and can also lead to erroneous detection.
As this is primarily a sensor limitation, this issue is hard to address from a mapping perspective.
Lastly, \rev{the approach relies on dense mapping of the scene, which requires sufficiently high data rates for the sensor speed}. Similarly, strong occlusions may impede complete mapping of new areas and thus lead to only partial detection at new map boundaries.

%due to the thin view frustum close to the sensor, objects starting close to the sensor may only be partially detected, as the free space is not yet sufficiently mapped. While this could be addressed with a range-based confidence, we find this to be rarely an issue in practice.

% ===============================================================================================

\section{Conclusions}
\label{sec:conclusion}
We presented \emph{Dynablox}, a novel online mapping-based approach for real-time moving object detection in complex dynamic environments. %, such as public places and work places. 
Benchmarks on real-world data achieve 86\% IoU at 17 FPS in typical robotic settings. 
The method outperforms a recent appearance-based classifier without making any assumptions on object or environment, and approaches the performance of a recent offline approach with hindsight knowledge.
We also qualitatively validated this generalization on a new custom data set with rare objects in challenging scenes.
Our experimental analysis of the computational scaling matched theoretical predictions, allowing real-time operation in typical indoor environments and adding only a 39\% overhead to a conventional volumetric mapping stack.
We additionally provide an easy formula for optimizing the method to the expected level of drift and validated this in a series of experiments. Finally, our implementation and data are released open-source to aid future research in the area. 

%Reiterate that this is an important problem, our claimed contributions, and how our experiments validated them. Laser focus.

% ===============================================================================================

\section*{ACKNOWLEDGMENT}
We thank Samuel Gull for support with the data collection.

% ===============================================================================================

{\small
\bibliographystyle{IEEEtran}
% \balance
\bibliography{IEEEfull,references}
}

%%%%%%%%%%%%%%%%%%%%%%%%%%%%%%%%%%%%%%%%%%%%%%%%%%%%%%%%%%%%%%%%%%%%%%%%%%%%%%%%

\end{document}